\documentclass{article}

\usepackage{microtype}
\usepackage{graphicx}
\usepackage{subcaption}
\usepackage{booktabs}
\usepackage{tabularx}
\usepackage{multirow}
\usepackage[table]{xcolor}
\usepackage{fvextra}
\usepackage{float}
\usepackage{enumitem}
\usepackage{caption}
\usepackage{wrapfig}

\usepackage{hyperref}

\usepackage[T1]{fontenc}

\usepackage{amsmath}
\usepackage{amssymb}
\usepackage{mathtools}
\usepackage{amsthm}

\usepackage{iclr2026_conference}

\usepackage[capitalize,noabbrev]{cleveref}

\theoremstyle{plain}

\theoremstyle{definition}

\theoremstyle{remark}

\iclrfinalcopy

\title{Learning to Repair Lean Proofs from Compiler Feedback}

\begin{document}

\author{
Yiran Wang \\
University of Washington \\
\And
Simon Chess$^\ast$ \\
University of Washington
\And
Daniel Lee$^\ast$ \\
University of Washington
\And
Siyuan Ge$^\ast$ \\
University of Washington
\And
Ajit Mallavarapu$^\ast$ \\
University of Washington
\And
Jarod Alper \\
University of Washington
\And
Vasily Ilin \\
University of Washington \\
\texttt{vilin@uw.edu}
}

\def\iclrEqualContribution{$^\ast$Equal contribution}

\maketitle

\begin{abstract}
As neural theorem provers become increasingly agentic, the ability to interpret and act on compiler feedback is critical. However, existing Lean datasets consist almost exclusively of correct proofs, offering little supervision for understanding and repairing failures. We study Lean proof repair as a supervised learning problem: given an erroneous proof and compiler feedback, predict both a corrected proof and a natural-language diagnosis grounded in the same feedback. We introduce APRIL (Automated Proof Repair in Lean), a dataset of 260,000 supervised tuples pairing systematically generated proof failures with compiler diagnostics and aligned repair and explanation targets. Training language models on APRIL substantially improves repair accuracy and feedback-conditioned reasoning; in our single-shot repair evaluation setting, a finetuned 4B-parameter model outperforms the strongest open-source baseline. We view diagnostic-conditioned supervision as a complementary training signal for feedback-using provers.
\end{abstract}

\section{Introduction}
Formal theorem proving offers a strict setting for studying machine learning: proofs must be written in a formal language and verified by a proof assistant such as Lean, leaving no ambiguity of correctness. This property allows learning systems to receive exact verification signals and has motivated a surge of recent work on learning-based automated theorem proving. Large Language Models (LLMs) have achieved substantial gains on formal benchmarks and even reached Olympiad-level performance \cite{lin2025goedel,seed2025seedprover,harmonic2025aristotle}.

Despite these advancements, the dominant objective in current systems remains end-to-end proof generation: given a formal statement, the goal is to produce a complete proof that verifies successfully. Training data therefore consists primarily of correct proofs. Evaluation is likewise based on whether a valid proof is eventually found. Failures encountered during generation are typically used to guide exploration or policy optimization, but are not treated as supervised learning targets. As such, models receive limited supervision for interpreting diagnostics, explaining what went wrong, or proposing targeted edits. Recent work has also shown that iterative proof refinement using compiler feedback is 32-128x more efficient than pass@k \cite{seed2025seedprover}. 

In contrast, human proof development is inherently iterative and error-driven. Proof engineers routinely write partial proofs, inspect compiler diagnostics, and incrementally revise the code until verification succeeds. Intermediate failures are fully expected in the development process and the same feedback is used both to decide what to change and to communicate why the change is correct. However, most publicly available proof corpora preserve only final proofs and omit failed attempts. Large-scale libraries such as mathlib provide extensive collections of formally verified theorems, but do not record intermediate error states encountered during development \cite{mathlib2020}. Datasets derived from natural language autoformalization pipelines, including Lean Workbook and Herald, provide aligned natural--formal pairs and synthetic proof data, but likewise consist almost entirely of valid formal artifacts rather than trajectories of failed attempts and subsequent repairs \cite{ying2024leanworkbook,gao2024herald}. As a result, models trained on these resources have little direct exposure to feedback-conditioned iteration, including both producing corrected code and articulating diagnoses grounded in compiler messages.

\begin{figure}[t]
  \centering
  \includegraphics[width=0.7\textwidth]{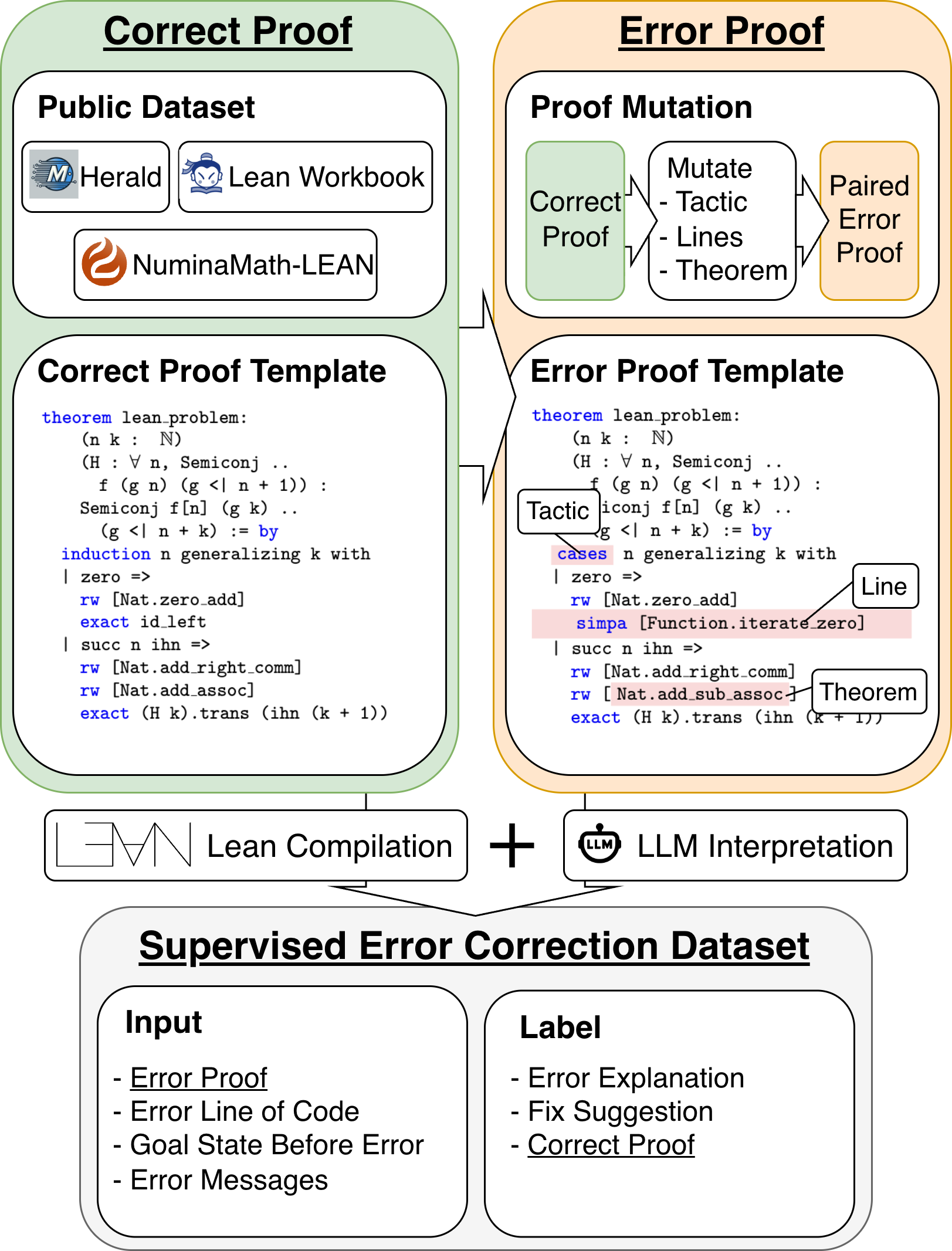}
  \caption{Overview of our dataset collection pipeline that collects paired correct and incorrect proof and their contextual information. We collected correct proofs from public datasets (Herald, Lean Workbook, NuminaMath-Lean). The paired error proofs that maintain the same proof sketch are generated by mutating tactics, lines of code, or theorems. Then, we used the Lean compiler to filter out the error proofs and extract their error messages, error lines, and goal states. Finally, we prompted an LLM to generate error explanations and fix suggestions based on the paired proofs and Lean Infoview.  The resulting dataset pairs erroneous proofs and error information with structured labels containing the error interpretation from LLM and the corresponding correct proofs.}
  \label{fig:03-03}
\end{figure}

This work studies proof correction from compiler feedback, and introduces a dataset that supports two aligned feedback-conditioned tasks: (1) producing a corrected proof, and (2) producing a natural-language diagnosis and fix suggestion grounded in the same feedback. These targets share the same underlying evidence (error message and local proof state), but emphasize different outputs: formal repair versus human-interpretable debugging. To enable controlled analysis, we construct datasets by systematically mutating correct proofs to generate plausible failures produced by controlled mutations, and we evaluate language models under supervised finetuning without reinforcement learning or inference-time search, isolating the model's ability to interpret diagnostics and apply targeted repairs.

Our contributions are threefold. First, we introduce a large-scale dataset of 260K Lean proof-repair examples, each consisting of an erroneous proof, compiler diagnostics including error messages and local proof state, and a corresponding corrected proof. Second, we develop a systematic mutation pipeline that generates realistic failures by substituting semantically related theorems, swapping similar tactics, and eliciting plausible but incorrect line and multi-line completions from language models. Third, we demonstrate that supervised finetuning on this data substantially improves repair accuracy under our evaluation setting: 27.4\% correction accuracy in our single-shot repair evaluation (no search/iteration), compared to 1.1\% for the base model and 26.8\% for Goedel-Prover-V2-32B under the same protocol. Because Goedel is typically deployed with search/iteration, this comparison should be interpreted as an ablation of compiler-feedback-conditioned repair rather than end-to-end proving performance.

\section{Related Work}

\paragraph{Neural Theorem Proving in Lean.}
Recent learning-based systems for Lean primarily target end-to-end proof generation, combining large language models with search, reinforcement learning, and synthetic data pipelines \cite{xin2024deepseekprover,ren2025deepseek,lin2025goedel,wang2025kimina}. Infrastructure such as LeanDojo has enabled systematic benchmarking and retrieval-augmented proving \cite{yang2023leandojo}, while Lean 4 provides the formal foundation for this work \cite{demoura2021lean4}. Evaluation across these systems is framed in terms of theorem-level success metrics such as pass@k on benchmarks like miniF2F \cite{zheng2021minif2f}.

\paragraph{Verifier Feedback and Proof Repair.}
Verifier feedback and errors play an important role in existing pipelines, but primarily as control signals for exploration and policy optimization rather than as supervised correction targets \cite{lin2025goedel,xin2024deepseekprover,harmonic2025aristotle,lample2022hypertree,xin2024deepseekproverv15,ren2025deepseek}. More recent refinement-based systems incorporate compiler diagnostics into prompts to regenerate proofs across multiple iterations, demonstrating that iterative repair can be substantially more efficient than independent sampling \cite{seed2025seedprover,zhou2025solving}. Our work studies proof correction from compiler feedback as a supervised learning problem, providing aligned repair and diagnosis targets grounded in the same compiler messages and local proof state.

\paragraph{Proof Datasets and Training Corpora.}
Large-scale libraries such as mathlib provide extensive collections of formally verified theorems, but do not record intermediate error states encountered during development \cite{mathlib2020}. Datasets derived from natural language autoformalization pipelines, including Lean Workbook, Herald, and NuminaMath, provide aligned natural--formal pairs and synthetic proof data, but likewise consist almost entirely of valid formal artifacts rather than trajectories of failed attempts and subsequent repairs \cite{ying2024leanworkbook,gao2024herald,li2024numinamath}. The scarcity of error trajectories in public corpora remains a key bottleneck for training models that can interpret and act on compiler diagnostics.

\paragraph{Program Repair from Diagnostic Feedback.}
In software engineering, program repair has been widely studied as a supervised or self-supervised learning problem over paired erroneous and corrected programs, including approaches that condition on compiler error messages to guide targeted edits \cite{gupta2017deepfix,yasunaga2020drrepair,berabi2021tfix}. Several lines of work also generate repair supervision at scale when paired error--fix data is scarce, using compilers or analyzers to validate synthetic errors and candidate fixes \cite{yasunaga2021bifi,wei2025swerl}. Our work applies a similar philosophy to the formal verification setting: we systematically mutate correct Lean proofs to generate plausible failures, then train models for feedback-conditioned repair.

\section{Methodology}
\subsection{Data Collection}

Each entry in APRIL consists of (i) an erroneous Lean proof that fails to compile, (ii) a corresponding fixed proof that is syntactically similar but compiles successfully, (iii) the compiler error messages and proof state, (iv) a natural-language explanation of the errors, and (v) a natural-language fix suggestion describing how to transform the erroneous proof into the corrected one.

Because large-scale corpora of human-generated Lean errors are scarce while correct proofs are widely available, we construct APRIL backward: beginning with a correct proof and systematically introducing errors into it. Although our failures are induced by controlled mutations, every example is anchored in the expected elaboration behavior: the error message, offending location, and local goal state are produced by the Lean compiler in a fixed environment, and the repair target is a verified proof in that same environment. We start from verified correct proofs sourced from the Herald, Lean Workbook, and Numina datasets, retaining only those that compile under Lean 4.22.0-rc4. We then introduce errors using the mutation strategies described below and keep only mutated proofs that trigger compiler errors. To retrieve compiler feedback and compilation rates, we use Lean-Interact to interact with the Lean REPL \cite{leaninteract}. For each retained example, we use DeepSeek-V3-0324 \cite{deepseekai2025deepseekv3} to generate an explanation of the compiler feedback and a corresponding fix suggestion.

\begin{figure}[t]
  \centering
  \begin{subfigure}[t]{0.48\textwidth}
    \centering
    \includegraphics[width=\linewidth]{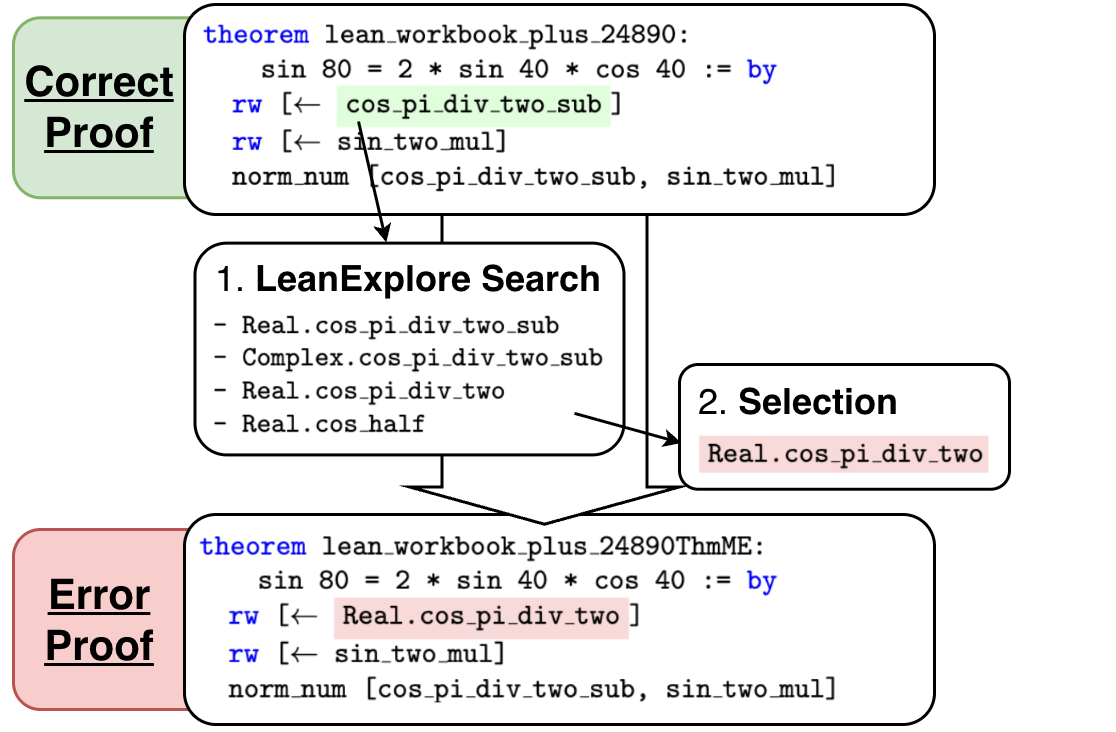}
    \caption{Theorem mutation error}
    \label{fig:theorem_mutation_error}
  \end{subfigure}
  \hfill
  \begin{subfigure}[t]{0.48\textwidth}
    \centering
    \includegraphics[width=\linewidth]{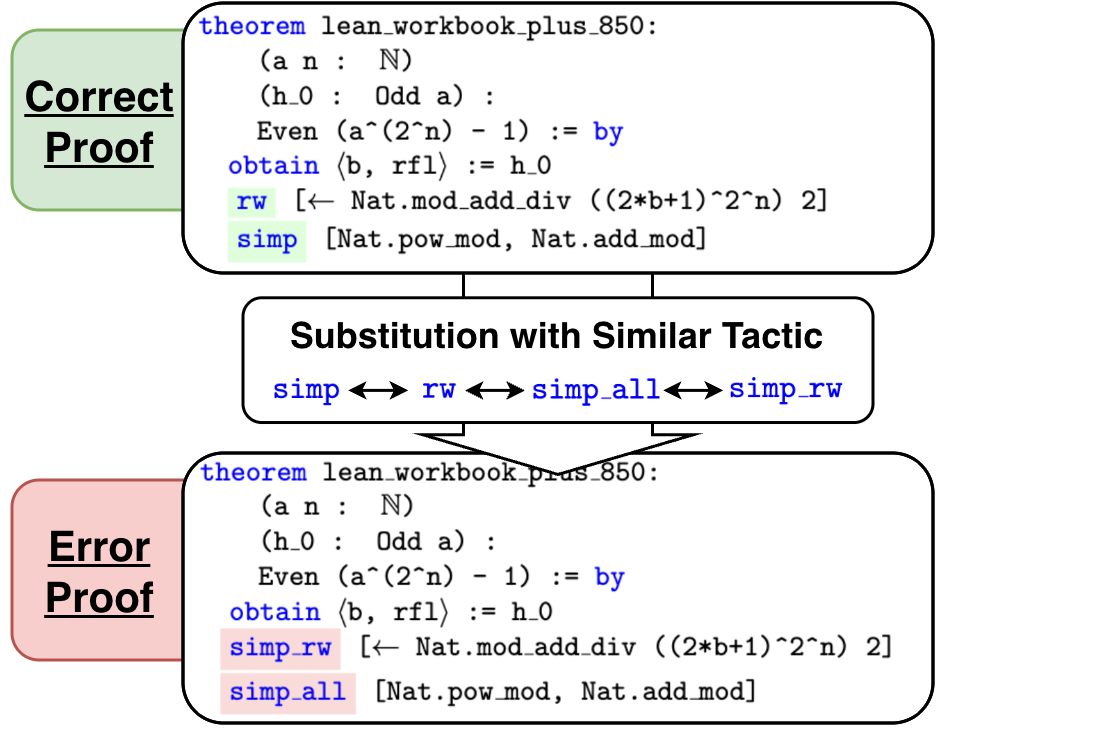}
    \caption{Tactic mutation error}
    \label{fig:tactic_mutation_error}
  \end{subfigure}

  \vspace{0.8em}

  \begin{subfigure}[t]{0.48\textwidth}
    \centering
    \includegraphics[width=\linewidth]{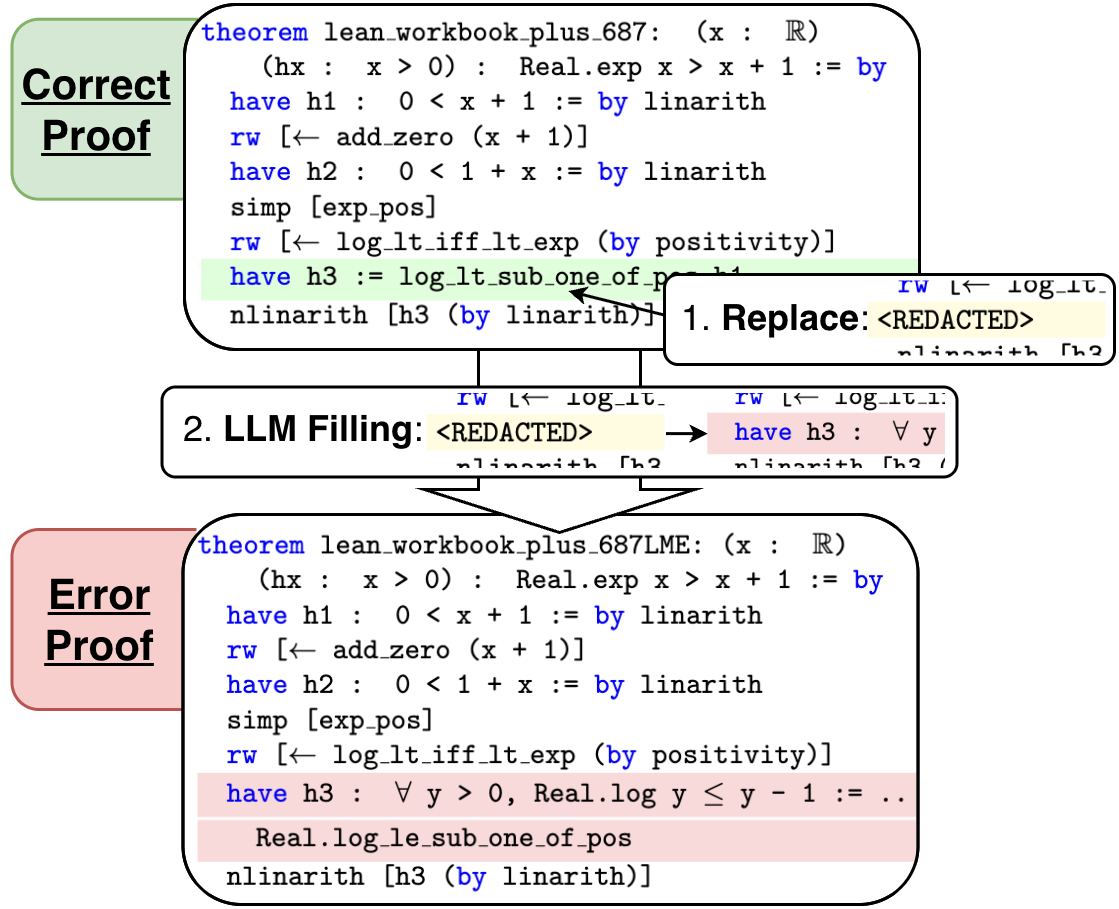}
    \caption{Line mutation error}
    \label{fig:line_mutation_error}
  \end{subfigure}
  \hfill
  \begin{subfigure}[t]{0.48\textwidth}
    \centering
    \includegraphics[width=\linewidth]{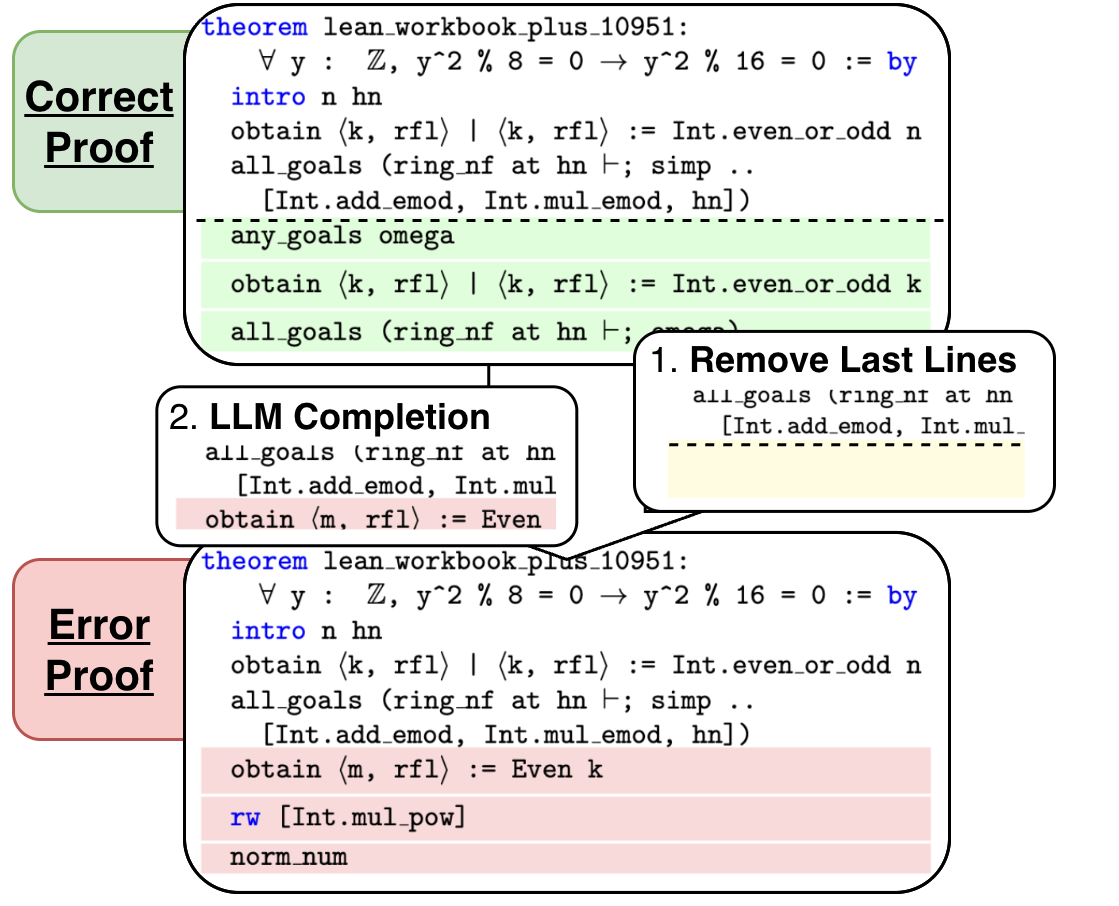}
    \caption{Multi-line mutation error}
    \label{fig:multiline_mutation_error}
  \end{subfigure}
  \caption{Examples of the four mutation error types with illustrations of their generation models.}
  \label{fig:mutation_errors}
\end{figure}

\paragraph{Theorem Mutation Errors}

Theorem mutation is designed to imitate errors that arise from subtle semantic mismatches, where a theorem is almost applicable but differs in its required premises or conclusion. To construct such mutations, we begin with proofs that compile successfully and extract the theorems used in each proof. Each resulting (proof, theorem) pair is treated as a separate mutation candidate. For each selected theorem occurrence, we retrieve semantically related declarations using the LeanExplore semantic search engine \cite{asher2025leanexplore}, filter out trivial replacements (including the original theorem and namespace-only variants), and perform a single substitution by replacing exactly one occurrence of the selected theorem identifier in the proof text. The remaining candidates typically differ in their hypotheses or conclusion in ways that are small but proof-critical, and the resulting error may surface later in the proof, far from the mutation site.

\paragraph{Tactic Mutation Errors}

Tactic mutations substitute a similar, yet incorrect tactic, e.g., swapping \texttt{nlinarith} with \texttt{linarith}, \texttt{norm\_num}, or \texttt{ring}. We define a fixed set of tactic equivalence classes based on common proof roles, including arithmetic solvers, rewriting tactics (e.g., \texttt{rw}, \texttt{simp}, \texttt{simp\_rw}), structural tactics (e.g., \texttt{intro}, \texttt{intros}, \texttt{rintro}), and proof-construction tactics (e.g., \texttt{apply}, \texttt{refine}, \texttt{exact}, \texttt{assumption}). Substitutions are only performed within the same class to ensure syntactic validity and plausibility. For each proof, we randomly select between one and three swappable tactic occurrences and substitute each with a similar alternative; we keep only unique mutated proofs that fail to compile.

\paragraph{Line Mutation Errors}

To produce a line mutation error, we begin with the corrected proof and at random replace one proof line (after the main \texttt{by}) with \texttt{REDACTED}, preserving indentation. This redacted proof is passed to DeepSeek-V3-0324 \cite{deepseekai2025deepseekv3} with instructions to provide the redacted line. The model-generated completion is reinserted and compiled with Lean; only mutated proofs that fail to compile are retained and duplicates are removed.

\paragraph{Multi-Line Mutation Errors}

Multi-line mutations are produced similarly to line mutations, except we redact the entirety of the proof after the randomly selected line and allow the model to produce as many lines as it pleases to replace the redacted lines. To ensure the erroneous proof resembles the correct proof, no more than half of the proof is redacted.

\paragraph{Explanation and Fix Generation}
All mutated proofs are compiled with Lean, and only those that produce a compiler error are retained, since some changes still lead to correct proofs. For each retained theorem mutation, we also store contrastive metadata, including the names and formal statements of both the intended (correct) theorem and the substituted (incorrect) theorem. For each retained erroneous proof, we automatically generate a natural language explanation of the failure and a suggested fix using a separate language model, prompted with the original proof, the mutated proof, the Lean compiler error message, and any available mutation metadata.

\subsection{Model finetuning}
We finetune Qwen3-4B-Instruct-2507 \cite{qwen3}, Kimina-Prover-Distill-8B, and Goedel-Prover-V2-8B using an identical supervised finetuning (SFT) pipeline with Low-Rank Adaptation (LoRA) \cite{hu2021lora}. All training procedures and hyperparameters are shared across all the experiments.

\paragraph{Training Format and Objective}
Each training example is converted into a chat-formatted prompt using the corresponding model tokenizer. The input consists of a system message and a user message concatenating the prover error, local proof state, and failing proof. The supervised target is the assistant completion, structured to encourage an explicit diagnostic reasoning step prior to code generation. The data format is as illustrated in Figure~\ref{finetuning-format} in Appendix~\ref{app:finetuning} .

\paragraph{LoRA Setup}
We apply LoRA adapters with rank 32 to both attention and MLP projection layers (\texttt{q\_proj}, \texttt{k\_proj}, \texttt{v\_proj}, \texttt{o\_proj}, \texttt{gate\_proj}, \texttt{up\_proj}, \texttt{down\_proj}) in each transformer block. This configuration is identical across all base models.

\paragraph{Optimization and Training Hyperparameters}
We use AdamW \citep{loshchilov2019adamw} with learning rate $1 \times 10^{-4}$, cosine decay, and linear warmup. Training runs for up to 15{,}000 steps with effective batch size 8 and maximum sequence length 2048. Early stopping with patience 5 (i.e., 1{,}250 steps) terminates training if validation loss does not improve, and we select the checkpoint with lowest validation loss. Full hyperparameter details are provided in Appendix~\ref{app:finetuning}.

\paragraph{Compute and Implementation Details}
All experiments are conducted on NVIDIA L40S and H200 GPUs using bfloat16 precision. We enable FlashAttention-2 \cite{dao2023flashattention2} when available, otherwise falling back to SDPA.

\section{APRIL}
\subsection{Dataset Composition}

\begin{table*}[ht]
\begin{center}
\begin{small}
\begin{sc}
\caption{Statistics of Correct Proof After Filtering. We report the number of raw and filtered proofs for three source datasets, along with the average proof length, average number of \texttt{have} statements per proof, and number of proofs containing at least one \texttt{have} statement. Across all datasets, we observe significant variation in proof length and \texttt{have} statement usage, with proofs sourced from human annotators in Numina's dataset demonstrating the most complexity in terms of length and \texttt{have} statements.}
\label{tab:data_source}
\begin{tabular}{lccccc}
\toprule
Dataset & Raw & Filtered & Avg. Lines & Avg. \texttt{have} & Contain \texttt{have} \\
\midrule
Herald & 30,190 & 16,010 & 4.02 & 0.13 & 1,500 \\
Lean Workbook & 10,433 & 9,491 & 2.71 & 0.16 & 820 \\
Numina Autoformalizer & 6,039 & 5,925 & 10.20 & 2.17 & 2,839 \\
Numina Human & 9,428 & 8,066 & 50.93 & 8.90 & 6,128 \\
\midrule
Total & 56,090 & 39,492 & 14.21 & 2.24 & 11,287 \\
\bottomrule
\end{tabular}
\end{sc}
\end{small}
\end{center}
\vskip -0.2in
\end{table*}

\paragraph{Correct Proof Sources.}
To construct paired correct and incorrect proofs that remain structurally aligned, we begin from large public corpora of Lean proofs and inject controlled errors into initially correct proofs. We collect correct proofs from the Herald, Lean Workbook, and NuminaMath-Lean datasets, and retain only samples that successfully compile under Lean~4.22.0-rc4.

In total, APRIL contains 39,492 unique compiled theorems with diverse proof styles and complexity (Table~\ref{tab:data_source}). Approximately 40.5\% of proofs originate from Herald, consisting of modular fragments extracted from intermediate states of human-written mathlib4 proofs. Another 24.0\% come from Lean Workbook and are short, low-complexity machine-generated proofs. The remaining proofs are drawn from NuminaMath-Lean: 15.0\% are produced by an autoformalizer and 20.4\% are annotated by human experts. These proofs exhibit longer average length and more frequent use of \texttt{have} statements, indicating higher structural complexity.

\begin{figure}[t]
  \centering
  \includegraphics[width=0.7\textwidth]{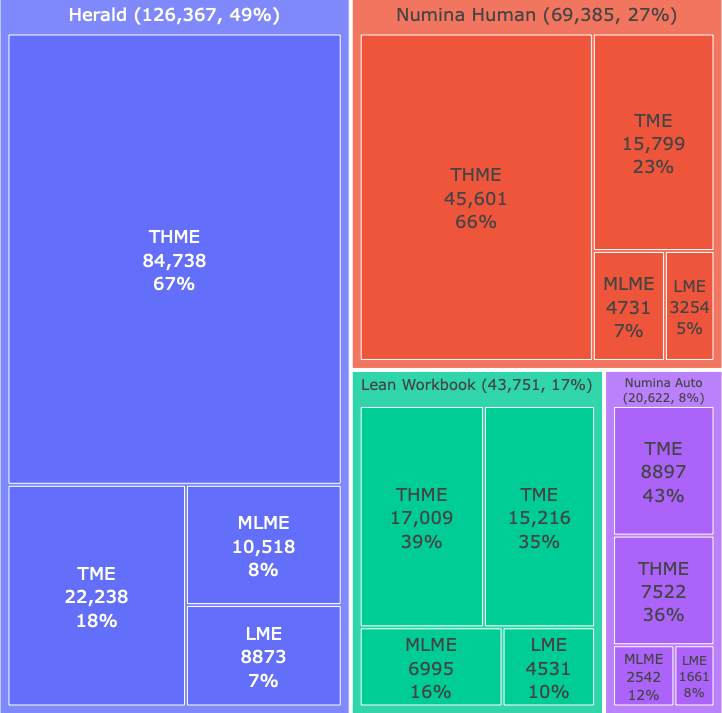}
  \caption{Statistics based on mutation type. For each dataset, we report the total number of erroneous proofs generated and detail the distribution of specific mutation types: Line Mutation Error (LME), Multi-line Mutation Error (MLME), Tactic Mutation Error (TME), and Theorem Mutation Error (THME). The percentages denote the proportion of each mutation type within its respective dataset.}
  \label{fig:detailed_stats}
\end{figure}

\paragraph{Generated Incorrect Proofs.}
From the 39,492 compiled theorems, we generate 260,125 incorrect proofs using four mutation operators (Figure~\ref{fig:detailed_stats}): (i) theorem substitution, (ii) tactic replacement, (iii) line-level redaction or modification, and (iv) multi-line redaction or modification. Each mutation yields a syntactically valid Lean file that fails to compile and produces a concrete compiler error trace.

Each dataset instance contains:
\begin{itemize}[noitemsep, topsep=0pt]
    \item the original correct proof,
    \item a mutated incorrect proof,
    \item the Lean compiler error message and local goal state at the failure location,
    \item a repaired proof target, and
    \item a natural-language explanation and fix suggestion aligned with the compiler feedback.
\end{itemize}

Theorem substitution errors constitute the largest fraction of incorrect proofs (59.5\%), reflecting the prevalence of type- and goal-mismatch failures in real proof development. See Appendix \ref{sec:attempts} for a detailed description of the discarded strategies for generating erroneous proofs.

\subsection{Dataset Split}

To prevent data leakage, we split the dataset at the level of original theorems rather than individual mutated proofs. All mutated variants derived from the same original theorem are assigned to the same split. This ensures that the model cannot observe identical correct proofs across training and evaluation. We additionally anonymize all theorem declarations by renaming them to a canonical identifier (lean\_problem), preventing models from exploiting cross-dataset name correlations or memorizing problem-specific identifiers.

We perform stratified splitting based on (i) source dataset (Herald, Lean Workbook, NuminaMath-Lean) and (ii) proof length, preserving the distribution of proof complexity and domains across splits. The proportions of each mutation type are also maintained in each split (Table~\ref{tab:dataset_splits}). Additional details on split statistics are provided in Appendix~\ref{app:more_data_description}.

\section{Results}

We evaluate proof repair accuracy on a held-out test set of 1{,}835 erroneous proofs spanning all four mutation types. A repair is considered successful if the model's output compiles under Lean 4.22.0-rc4. We compare against the base Qwen3-4B-Instruct model (no finetuning) and Goedel-Prover (8B/32B), using the same single-shot repair interface (no search/iteration).

\subsection{Main Results}

Table~\ref{tab:main-results} reports proof repair accuracy across models and error types. Finetuning on APRIL yields large gains across all models. For Qwen3-4B-Instruct, finetuning increases repair accuracy from 1.1\% to 27.4\% (25$\times$). Under the same protocol, this slightly exceeds Goedel-Prover-V2-32B (26.8\%). Finetuned 8B models reach 31--35\% repair accuracy, outperforming the 32B Goedel baseline.

All results are single-shot repairs from a failing attempt, not theorem-level success under multi-sample search or exploration.

\begin{table*}[h]
\caption{Proof repair accuracy by model and error type when training jointly for repair and explanation. Results outside the parentheses correspond to models jointly finetuned on all error types and evaluated in each category. Parenthesized values denote models finetuned separately on a single error type and evaluated on that same type. Section~5.3 analyzes the effect of removing explanations.}
\label{tab:main-results}
\vskip 0.15in
\begin{center}
\begin{small}
\begin{sc}
\begin{tabular}{lccccc}
\toprule
Baseline Model & Full & Tactic & Line & Theorem & Multi-Line \\
\midrule
Goedel-Prover-V2-8B
  & 15.5\% & 19.6\% & 20.0\% & 12.7\% & 19.4\% \\
Goedel-Prover-V2-32B
  & 26.8\% & 34.2\% & 28.5\% & 23.0\% & 32.6\% \\
Kimina-Prover-1.7B
  & 8.4\% & 15.1\% & 13.5\% & 4.8\% & 10.4\% \\
Kimina-Prover-8B
  & 11.1\% & 17.3\% & 14.5\% & 7.9\% & 13.9\% \\
Qwen3-4B-Instruct-2507
  & 1.1\% & 1.8\% & 2.5\% & 0.5\% & 0.0\% \\
\midrule
Finetuned Goedel-8B
  & 34.6\% & 41.7\% & 18.5\% & 36.8\% & 20.8\% \\
Finetuned Kimina-8B
  & 31.9\% & 38.9\% & 18.5\% & 34.1\% & 14.6\% \\
Finetuned Qwen3-4B
  & 27.4\% & 39.7\% & 16.0\% & 26.8\% & 13.2\% \\
\bottomrule
\end{tabular}

\end{sc}
\end{small}
\end{center}
\vskip -0.1in
\end{table*}



\subsection{Analysis by Error Type}

Table~\ref{tab:main-results} further breaks down repair performance by mutation category, revealing systematic variation in difficulty across error types while preserving consistent trends across models. Tactic mutations yield the highest repair rates, with top performance reaching 42.5\%, reflecting the relatively local nature of these errors within a fixed proof context. Theorem mutations exhibit intermediate difficulty, while line mutations are the most challenging, with maximum accuracy of 13.5\%, as they often involve correcting semantically inconsistent or hallucinated proof steps.

Importantly, models trained jointly on the full dataset maintain strong performance when evaluated on individual error categories. For example, Qwen3-4B-Instruct fine-tuned on the full dataset achieves competitive accuracy across tactic, theorem, and multi-line errors without explicit specialization. Comparable behavior is observed for Goedel-Prover-8B and Kimina-Prover-8B, where training on the full dataset results in only modest differences relative to models trained separately on each error type.

The limited degradation observed under joint training indicates that the dataset’s mutation types share substantial structural overlap. Rather than inducing negative interference, joint training enables models to learn repair strategies that generalize across error categories. This property is particularly important for realistic proof repair settings, where error types are heterogeneous and not known a priority.

\subsection{Effect of Explanations}

We ablate the effect of jointly supervising natural-language explanations alongside proof repair. Specializing exclusively on repair increases pass@1 from 27.4\% to 31.2\%. This reveals a controllable trade-off: training exclusively for repair maximizes autonomous pass@1, while joint supervision yields human-interpretable diagnoses that can support human-in-the-loop debugging or downstream tool-using agents. 

The explanations that finetuned Qwen produce are valuable in their own right, however. Model explainability is often hard to come by, and this is especially true for error correction in Lean. Top models (Goedel and Kimina) are specialized in error correction to the point that they have lost much of their ability to produce meaningful natural language, including to explain their work, making their reasoning very difficult to understand. Training on our dataset allows Qwen to improve significantly at correcting Lean errors and at the cost of an only slight reduction in error correction accuracy it can also produce explanations of its reasoning. As a small demonstration of the downstream utility of explanations, we provide DeepSeek with the same failing instance augmented by an explanation, and observe substantially higher success rates when using explanations produced by the explanation-trained model. DeepSeek succeeds with a rate of 4\% when aided by base Qwen's explanations and 29\% when aided by the trained model's explanations, performing better than either model individually.

\begin{table}[ht]
\centering

\label{tab:results}
\vskip 0.15in
\begin{tabular}{l l c c c}
\toprule
Model & Method & w/o exp & w/ exp & $\Delta$ \\
\midrule
\multirow{5}{*}{Goedel-8B}
 & Full        & 36.7\% & 34.6\% & \cellcolor{red!15}\makebox[2.5em]{-2.1\%} \\
 & Tactic      & 48.5\% & 46.5\% & \cellcolor{red!10}\makebox[2.5em]{-2.0\%} \\
 & Line        & 25.5\% & 21.5\% & \cellcolor{red!20}\makebox[2.5em]{-4.0\%} \\
 & Theorem     & 37.5\% & 37.0\% & \cellcolor{red!5}\makebox[2.5em]{-0.5\%} \\
 & Multi-Line  & 24.3\% & 22.9\% & \cellcolor{red!10}\makebox[2.5em]{-1.4\%} \\
\midrule
\multirow{5}{*}{Kimina-8B}
 & Full        & 36.9\% & 31.9\% & \cellcolor{red!25}\makebox[2.5em]{-5.0\%} \\
 & Tactic      & 49.7\% & 47.2\% & \cellcolor{red!15}\makebox[2.5em]{-2.5\%} \\
 & Line        & 23.5\% & 23.5\% & \cellcolor{gray!10}\makebox[2.5em]{0.0\%} \\
 & Theorem     & 37.4\% & 37.0\% & \cellcolor{red!5}\makebox[2.5em]{-0.4\%} \\
 & Multi-Line  & 26.4\% & 21.5\% & \cellcolor{red!20}\makebox[2.5em]{-4.9\%} \\
\bottomrule
\end{tabular}
\caption{Pass@1 on error types matching the training regime. $\Delta$ denotes the change from w/o exp (color-coded).}
\end{table}

\section{Conclusion}
We introduced Lean proof repair as a supervised learning task: given a failing proof and compiler feedback, predict a corrected proof that compiles. To support this, we constructed APRIL, a dataset of paired erroneous and verified proofs augmented with compiler diagnostics and additional natural-language annotations. Supervised finetuning on this data substantially improves repair accuracy, with a finetuned 4B model outperforming larger open-source provers under our evaluation setup. These results suggest that error-centric supervision is a strong and underutilized training signal for building agentic theorem provers that can iteratively refine proofs from feedback.

\section*{Reproducibility}
The APRIL dataset is available at \url{https://huggingface.co/datasets/uw-math-ai/APRIL}. Finetuned models are available at \url{https://huggingface.co/uw-math-ai/gAPRIL-w-exp} (with explanations) and \url{https://huggingface.co/uw-math-ai/gAPRIL-wo-exp} (without explanations).

\section*{Acknowledgements}
CPU and GPU computing, LLM inference and data storage were in part done using AWS credits from the UW eScience School and UW IT. CPU and GPU computing was in part done using the UW Research Computing Club funded from the UW Student Technology Fee Committee. GPU computing was in part done on UWIT's GPU cluster Tillicum. GPU computing and model inference were in part done using Nebius and TokenFactory.
\newpage

\bibliographystyle{iclr2026_conference}
\bibliography{refs}

\newpage
\appendix
\onecolumn
\section{Dataset Description}
\label{app:more_data_description}

\begin{table}[ht]
\centering
\caption{Number of erroneous proofs by split}
\label{tab:dataset_splits}
\vskip 0.15in
\begin{small}
\begin{sc}
\begin{tabular}{lcccc}
\toprule
Error Type & Train & Val & Test & Total \\
\midrule
Full       & 249{,}027 & 9{,}263 & 1{,}835 & 260{,}125\\
\midrule
Tactic     & 59{,}688 & 2{,}064 & 398 & 62,150\\
Line       & 17{,}098  & 1{,}021 & 200 & 18,319\\
Theorem    & 148{,}362 & 5{,}415 & 1{,}093 & 154,870\\
Multi-Line & 23{,}879  & 763     & 144 & 24,786\\
\bottomrule
\end{tabular}
\end{sc}
\end{small}
\end{table}

\begin{table*}[ht]
\centering
\caption{Basic dataset statistics by split.}
\label{tab:dataset_basic_stats}
\vskip 0.15in
\small
\scshape
\begin{tabular}{lcccc}
\toprule
Split & Num & Avg. Lines & Avg. have & Contain have \\
\midrule
Train & 38,292 & 14.22 & 2.25 & 10,966 (28.64\%) \\
Val   & 1,000  & 13.74 & 1.85 & 267 (26.70\%) \\
Test  & 200    & 14.29 & 1.71 & 54 (27.00\%) \\
\midrule
Total & 39,492 & 14.21 & 2.24 & 11,287 (28.58\%) \\
\bottomrule
\end{tabular}
\end{table*}

\begin{table*}[ht]
\centering
\caption{Statement source distribution by split.}
\label{tab:dataset_source_stats}
\vskip 0.15in
\small
\scshape
\begin{tabular}{lcccc}
\toprule
Split & Lean Workbook & Herald & Numina Auto & Numina Human \\
\midrule
Train & 9,159 (23.92\%) & 15,471 (40.40\%) & 5,833 (15.23\%) & 7,829 (20.45\%) \\
Val   & 276 (27.60\%)   & 450 (45.00\%)    & 77 (7.70\%)     & 197 (19.70\%) \\
Test  & 56 (28.00\%)    & 89 (44.50\%)     & 15 (7.50\%)     & 40 (20.00\%) \\
\midrule
Total & 9,491 (24.03\%) & 16,010 (40.54\%) & 5,925 (15.00\%) & 8,066 (20.42\%) \\
\bottomrule
\end{tabular}
\end{table*}

\section{Finetuning Setting}
\label{app:finetuning}

\begin{figure}[H]
\vskip -0.2in
\begin{center}
\centerline{\includegraphics[width=0.5\columnwidth]{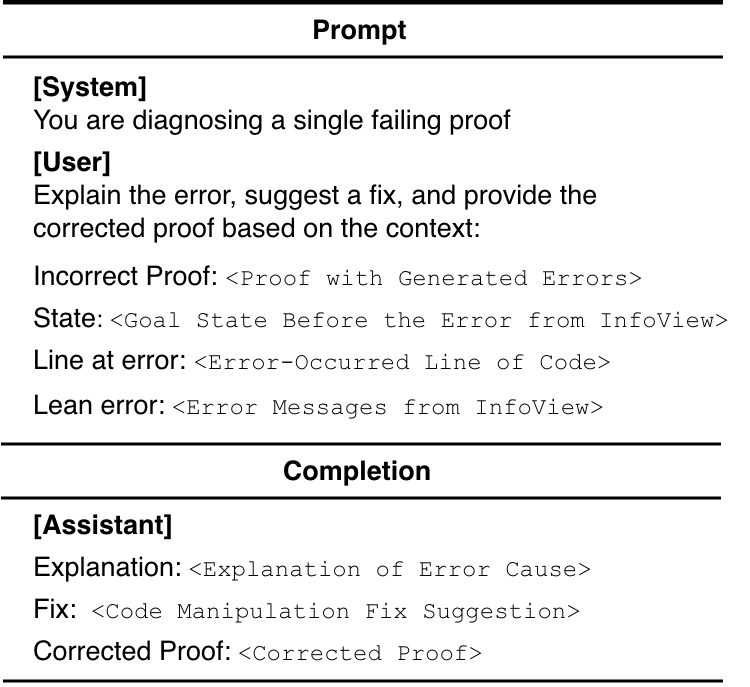}}
\caption{A template of our finetuning data format}
\label{finetuning-format}
\end{center}
\vskip -0.2in
\end{figure}

\begin{table}[h]
\centering
\caption{Optimization and training hyperparameters used in all experiments.}
\label{tab:training-hparams}
\vskip 0.15in
\begin{small}
\begin{tabular}{ll}
\toprule
\textbf{Hyperparameter} & \textbf{Value} \\
\midrule
Evaluation metric & \texttt{eval\_loss} \\
Effective batch size & 8 \\
Optimizer & AdamW \\
Learning rate & $1 \times 10^{-4}$ \\
LR scheduler & Cosine decay \\
Warmup ratio & 0.1 \\
Weight decay & 0.01 \\
Max gradient norm & 1.0 \\
Sequence length & 2048 \\
Sequence packing & Disabled \\
Precision & bfloat16 (bf16) \\
Evaluation frequency & Every 250 steps \\
Checkpoint frequency & Every 250 steps \\
Checkpoint selection & Best validation loss \\
\midrule
Rank ($r$) & 32 \\
Scaling factor ($\alpha$) & 64 \\
Dropout & 0.1 \\
Bias & None \\
Target modules & \texttt{q\_proj, k\_proj, v\_proj,} \\
 & \texttt{o\_proj, gate\_proj,} \\
 & \texttt{up\_proj, down\_proj} \\
\bottomrule
\end{tabular}
\end{small}
\end{table}

\section{Prompts for Explanation and Fix Suggestion Generation}

DeepSeek is involved in the generation of each type of error. In the case of theorem and tactic mutations, we only use it for generating the explanation and fix suggestion. In the case of the line and multiline mutations, we also use it to produce the errors themselves. We provide our prompts for each of these protocols below.

\subsection{Line Mutation Errors}
\paragraph{Error generation system prompt:}
\begin{Verbatim}
You are a Lean 4 programmer.
\end{Verbatim}
\paragraph{Error generation instruction block:} 
\begin{Verbatim}[
  breaklines=true,
  breakanywhere=true,
  breaksymbolleft={},
  breaksymbolright={}
]
One line has been redacted in this lean4 proof. Please complete the proof by providing the correct contents of the redacted line. Your response will be automatically searched for your answer. To facilitate this, please write "MY ANSWER" before your answer. Your answer should be exactly one line long and should contain no semicolons. For example, if you were given
```lean4
theorem very_simple: 1+1=2 := by
  REDACTED
```
you might respond with
"""
This is very easy, `rfl` accomplishes this in Lean 4.
MY ANSWER
```lean4
rfl
```
"""
Now try this theorem
```lean4
{broken\_proof}
```
\end{Verbatim}
\paragraph{Explanation system prompt}
\begin{Verbatim}[
  breaklines=true,
  breakanywhere=true,
  breaksymbolleft={},
  breaksymbolright={}
]
You are a Lean 4 programmer diagnosing one failing proof. Assume you ONLY see the incorrect proof text, the infoview state near the failure, and Lean's error message.
\end{Verbatim}

\paragraph{Explanation instruction block.}
\begin{Verbatim}[
  breaklines=true,
  breakanywhere=true,
  breaksymbolleft={},
  breaksymbolright={}
]
Explain why the incorrect proof fails and how to correct it using only the incorrect proof, infoview state, and error.

Return ONLY one JSON object with exactly these two fields:
{
    "explanation": "1--3 sentences explaining the concrete reason the proof fails",
    "fix_suggestion": "1 sentence with a high-level fix (no code);
}
No code blocks. No extra fields. Both fields must be non-empty.
\end{Verbatim}
\subsection{Theorem Mutation Errors}

\paragraph{System prompt.}
\begin{Verbatim}[
  breaklines=true,
  breakanywhere=true,
  breaksymbolleft={},
  breaksymbolright={}
]
You are a Lean 4 programmer diagnosing one failing proof.
You will see the incorrect proof, its state/error, and a cheatsheet of metadata detailing the intended (correct) theorem versus the one that was substituted (incorrect).
Use this metadata to explain the failure.
\end{Verbatim}

\paragraph{Instruction block.}
\begin{Verbatim}[
  breaklines=true,
  breakanywhere=true,
  breaksymbolleft={},
  breaksymbolright={}
]
Explain why the proof fails by contrasting the incorrect vs intended theorem.

Return ONLY one JSON object with exactly these two fields:
{
    explanation: 1--3 sentences explaining the concrete reason the proof fails,
    fix_suggestion: Start with: Replace (incorrect_name) with (correct_name), and briefly say why that resolves the mismatch.
}
No code blocks. No extra fields. Both fields must be non-empty.
\end{Verbatim}

\subsection{Tactic Mutation Errors}
\paragraph{Explanation system prompt}
\begin{Verbatim}[
  breaklines=true,
  breakanywhere=true,
  breaksymbolleft={},
  breaksymbolright={}
]
You are a Lean 4 programmer diagnosing one failing proof.
You will see an incorrect proof containing one or more invalid tactics. You will also see its state/error and a 'cheatsheet' of metadata detailing the intended (correct) line versus the current (incorrect) line containing a swapped tactic.
The proof may contain multiple independent tactic failures. The compiler error may only reflect the first encountered failure. Your explanation and fix suggestion should consider all incorrect tactics shown.
Use this metadata to explain the failure.

\end{Verbatim}
\paragraph{Instruction block.}
\begin{Verbatim}[
  breaklines=true,
  breakanywhere=true,
  breaksymbolleft={},
  breaksymbolright={}
]
Explain why the proof fails by contrasting the incorrect line vs the intended line.

Return ONLY one JSON object with exactly these two fields:
{
    explanation: 1-3 sentences explaining the concrete reason why the applied tactic(s) fail to make progress on the goal,including reasoning about goal structure, type, or required properties, without directly mentioning the replacement tactic.,
    fix\_suggestion: "Start with EXACTLY the following format: 'Replace `FULL_INCORRECT_TACTIC` with `FULL_INTENDED_TACTIC` on Line X because EXPLANATION'. Use the full tactic call including arguments. If multiple errors exist, list fixes for all.
}
No code blocks. No extra fields. Both fields must be non-empty.

\end{Verbatim}

\section{Unsuccessful Attempts for Data Synthesis}\label{sec:attempts}
In addition to the mutation strategies described above, we explored several alternative approaches for generating realistic erroneous proofs. These methods were motivated by the goal of producing more diverse and challenging failures, but in practice they exhibited limitations in controllability, fidelity, or scalability. We briefly summarize these unsuccessful attempts below to clarify the design choices that led to our final dataset construction pipeline.

\paragraph{Introduce an (interesting) error.}
One approach we explored was directly prompting a language model to introduce an "interesting" error into an otherwise correct proof. In practice, this strategy exhibited very low diversity: even when increasing sampling temperature, the model tended to produce the same small set of superficial     modifications. The resulting errors often did not meaningfully stress semantic reasoning about the proof state.

\paragraph{Single-line translation to natural language.}
We additionally explored translation-based error generation by "translating" an individual Lean proof into natural language and then converting them back into Lean. Since formal logic is removed when converting to natural language, we hoped that this approach would produce realistic, interesting errors. However, we found that single-line translations struggled without the context of the theorem, resulting in errors that were unrealistic in the context of the theorem.

\paragraph{Full-proof translation to natural language.}
Since single-line translation performed poorly due to a lack of context, we repeated this translation process with the entire proof. For this approach, we focused solely on the proof body and separated it from the theorem statement. When converting from natural language back to Lean, we included the formalized theorem header to provide context. Despite these changes, the output proof was deemed too dissimilar from the original proof to be considered useful for error correction.

\paragraph{Full-proof translation to alternative proof assistants.}
One method we attempted for producing interesting errors was having an LLM translate the entire proof into another proof assistant and then back to Lean 4. We attempted this with both Rocq \cite{rocq900} and Lean 3 but were unsuccessful in both cases. When translating to Rocq and back, the incorrect proof that was produced would be too dissimilar from the correct proof to be considered an erroneous version of it, making it unhelpful for training. In the case of Lean 3, the proofs would often be similar, but the proof would often compile even after the translation. In the cases when it did not compile, the errors rarely resembled those that would actually be made in a typical effort to prove the theorem, often being overly simplistic. 

\paragraph{Common Lean pitfalls.}
We explored synthetic data generation by prompting a large language model to introduce single, controlled Lean proof errors drawn from well-known Lean pitfalls (e.g., misuse of have for data extraction, rewriting under binders, or confusing b > a with a < b). Despite providing explicit pitfall categories and constraining the model to preserve the original statement structure, the generated outputs were largely unusable. The model frequently ignored the specified pitfall, instead modifying unrelated parts of the proof, altering the theorem statement itself, or introducing multiple cascading errors rather than a single localized failure. In many cases, the modified proofs still compiled, indicating that the intended error was not semantically realized in Lean. When errors did occur, they often reflected generic type mismatches or syntax issues rather than the targeted pitfall (e.g., confusing Prop and Bool or mishandling inequalities). Manual inspection showed that the model tended to delete or rewrite 5–10 lines arbitrarily, sometimes removing entire proof segments, and lacked sensitivity to Lean’s proof-state evolution and elaboration constraints.
\paragraph{Random Multi-line redactions.}
We attempted to produce a type of error similar to line mutations in which a random number of consecutive lines at a random location were redacted and a model was prompted to fill them in. This was unsuccessful because the models would fail to take the Lean code that followed their section into account, resulting in "no goals to solve" errors, as well as problems with indentation and other syntactical frustrations. This occurred even when using more powerful models such as Gemini 3 \cite{google2025gemini3}.

\paragraph{Proof Repair via Prover-Based Pipelines.}
We evaluated a two-stage proof generation and repair pipeline. First, we prompted Kimina-Prover-Distill-1.7B \cite{wang2025kimina} to provide proofs for theorems in the dataset. The proofs that failed to verify were then passed to Goedel-Prover-V2-8B. This approach rarely produced valid proof repairs: Goedel-Prover typically rewrote the proof from scratch, substantially altering the proof structure and reasoning. As a result, the outputs could not reasonably be characterized as corrections of the original incorrect proofs, but rather as independent re-proofs.
\section{Data Synthesis Expanded}
\paragraph{Theorem Mutation Errors}

Theorem mutation is designed to imitate errors that arise from subtle semantic mismatches. The primary motivation is to force models to reason precisely about types, hypotheses, and goal structure, since many failures in real proof development occur when a theorem is almost applicable but differs in its required premises or conclusion. By substituting one valid theorem with another that is semantically related but type-incompatible in the current context, we generate failures that require understanding why a particular statement does not fit the goal, rather than merely detecting malformed code.

To construct such mutations, we begin with proofs that are known to compile successfully. For each proof, we extract the theorems used in the proof. Each resulting (proof, theorem) pair is treated as a separate mutation candidate, allowing a single proof to yield multiple independent mutation opportunities.

For each selected theorem occurrence, we retrieve semantically related declarations using the LeanExplore semantic search engine, queried through its Python API \cite{asher2025leanexplore}. LeanExplore indexes Lean 4 declarations using a combination of symbolic features and learned embeddings, enabling retrieval of theorems that are conceptually related even when their names or namespaces differ. From the retrieved candidates, we filter out trivial replacements, including the original theorem itself and declarations that differ only in the namespace. The remaining candidates typically differ in their hypotheses or conclusion in ways that are small but proof-critical.

For each mutation candidate, we retrieve up to 5 nearest semantic neighbors from LeanExplore and randomly sample from the filtered candidate set to produce a single substitution per candidate. In practice, a single proof can yield multiple independent theorem mutations corresponding to different theorem occurrences. Mutation is performed by replacing exactly one occurrence of the selected theorem identifier in the proof text with a retrieved alternative, leaving the rest of the proof unchanged.

This design encourages errors that will sometimes propagate through later steps of the proof, often surfacing at points far from the mutation site, similar to failures encountered during real interactive development.

\paragraph{Tactic Mutation Errors}

Similar to theorem mutations, tactic mutations involve a substitution of a similar, yet incorrect tactic. This approach requires grouping tactics that serve similar roles, such as arithmetic solvers or rewriting tactics, to ensure that a substitution is syntactically valid and plausible. For example, an arithmetic-solving tactic such as \texttt{nlinarith} may be substituted with \texttt{linarith}, \texttt{norm\_num}, or \texttt{ring}, which are syntactically valid but likely to fail in its current proof state.

We define a fixed set of tactic equivalence classes based on common proof roles, including arithmetic solvers (e.g., \texttt{linarith}, \texttt{nlinarith}, \texttt{norm\_num}, \texttt{ring}), rewriting tactics (e.g., \texttt{rw}, \texttt{simp}, \texttt{simp\_rw}), structural tactics (e.g., \texttt{intro}, \texttt{intros}, \texttt{rintro}), and proof-construction tactics (e.g., \texttt{apply}, \texttt{refine}, \texttt{exact}, \texttt{assumption}). Substitutions are only performed within the same equivalence class to ensure syntactic validity and plausibility.

For each proof that is known to compile successfully, we identify each occurrence of a swappable tactic and randomly select between one and three tactic occurrences. Each selected tactic is substituted with a similar alternative. With each successful tactic mutation, metadata, specifically the original line, substituted line, and the line number, are included to provide additional context for error explanation. Only unique proofs that failed to compile are kept.

\paragraph{Line Mutation Errors}

To produce a line mutation error, we begin with the corrected proof and at random replace one of its proof lines (a line occurring after the main \texttt{by}) with \texttt{REDACTED}, preserving its original indentation. This redacted proof is then passed to DeepSeek-V3-0324 \cite{deepseekai2025deepseekv3} with instructions to provide the redacted line that will cause the code to compile. By instructing the model to produce accurate code,


\end{document}